\documentclass[conference]{IEEEtran}
\IEEEoverridecommandlockouts
\usepackage{cite}
\usepackage{amsmath,amssymb,amsfonts}
\usepackage{algorithmic}
\usepackage{graphicx}
\usepackage{textcomp}
\usepackage{xcolor}
\usepackage{algorithm}
\def\BibTeX{{\rm B\kern-.05em{\sc i\kern-.025em b}\kern-.08em
    T\kern-.1667em\lower.7ex\hbox{E}\kern-.125emX}}

\def\argmax{\mathop{\rm argmax}}
\def\argmin{\mathop{\rm argmin}}
\def\diag{\mathop{\rm diag}}

\def\RR{\mathbb R}

\def\N{{\cal N}}

\def\N{{\cal N}}
\def\RR{\mathbb R}

\def\N{{\cal N}}

\def\N{{\cal N}}
\newcommand{\bmu}{{\boldsymbol{\mu}}}

\newcommand{\bbeta}{\mathbf{\beta}}

\newcommand{\bb}{\mathbf{b}}
\newcommand{\btheta}{{\boldsymbol{\theta}}}

\newcommand{\bsigma}{{\boldsymbol{\sigma}}}
\newcommand{\bSigma}{{\boldsymbol{\Sigma}}}

\newcommand{\be}{\mathbf{e}}
\newcommand{\bh}{\mathbf{h}}

\newcommand{\bx}{\mathbf{x}}

\newcommand{\bu}{\mathbf{u}}

\newcommand{\bz}{\mathbf{z}}
\graphicspath{{figures/}}
\DeclareGraphicsExtensions{.jpg,.pdf,.mps,.png}

\begin{document}

\title{Joint Representation Learning and Clustering via Gradient-Based Manifold Optimization}

\author{
    \IEEEauthorblockN{Sida Liu}
    \IEEEauthorblockA{\textit{Department of Statistics} \\
    \textit{Florida State University}\\
    Tallahassee, Florida, USA \\
    sl15r@my.fsu.edu}
    \and
    \IEEEauthorblockN{Yangzi Guo}
    \IEEEauthorblockA{\textit{Department of Mathematics} \\
    \textit{Florida State University}\\
    Tallahassee, Florida, USA \\
    yg12@my.fsu.edu}
    \and
    \IEEEauthorblockN{Mingyuan Wang}
    \IEEEauthorblockA{\textit{Department of Statistics} \\
    \textit{Florida State University}\\
    Tallahassee, Florida, USA \\
    mw15m@my.fsu.edu}
}

\maketitle

\begin{abstract}
Clustering and dimensionality reduction have been crucial topics in machine learning and computer vision. 
Clustering high-dimensional data has been challenging for a long time due to the curse of dimensionality. 
For that reason, a more promising direction is the joint learning of dimension reduction and clustering. 
In this work, we propose a Manifold Learning Framework that learns dimensionality reduction and clustering simultaneously. 
The proposed framework is able to jointly learn the parameters of a dimension reduction technique (e.g. linear projection or a neural network) and cluster the data based on the resulting features (e.g. under a Gaussian Mixture Model framework). 
The framework searches for the dimension reduction parameters and the optimal clusters by traversing a manifold,using Gradient Manifold Optimization. The obtained 
The proposed framework is exemplified with a Gaussian Mixture Model as one simple but efficient example, in a process that is somehow similar to unsupervised Linear Discriminant Analysis (LDA).
We apply the proposed method to the unsupervised training of simulated data as well as a benchmark image dataset (i.e. MNIST). The experimental results indicate that our algorithm has better performance than popular clustering algorithms from the literature.
\end{abstract}

\begin{IEEEkeywords}
Unsupervised Learning, Manifold Learning, style, styling, insert
\end{IEEEkeywords}

\section{Introduction}
\label{sec:intro}
In  recent years, unsupervised learning algorithms have become the building blocks in many machine learning and computer vision applications.

Clustering and dimensionality reduction, as two major topics in unsupervised learning, have been widely applied in various applications of computer vision and machine learning \cite{shi2000normalized, joulin2010discriminative, hershey2016deep}, including but not limited to feature matching, image segmentation and separation, etc. 

In this work, we propose a novel framework for simultaneously learning clustering assignments and dimensionality reduction. 
Our experiments demonstrate that it is feasible to seek for a projection of high dimensional data on a low-dimensional space that can be adequately separable to perform clustering. 
The framework we propose in this work is general and is not restricted to a specific task and scenario in machine learning and computer vision. 
We also illustrate that a simple implementation of our proposed framework with a linear projection or a shallow feed-forward Neural Network (FNN) as dimension reduction and isotropic Gaussian Mixture Model as the model-based clustering scheme can be sufficient for some clustering tasks. Furthermore, our framework can potentially be adapted to other machine learning tasks on large scale datasets based on state-of-art deep Convolutional Neural Networks.

The framework can generally be regarded as an unsupervised version of Linear Discriminant Analysis (LDA) \cite{izenman2013linear}. With certain feature representations learned by dimensionality reduction, the objective of the proposed framework and algorithm is to maximize clustering performance by a clustering evaluation function, where the clustering parameters are obtained by minimizing a clustering criterion that measures how the model fits the data. 
The novelty of our algorithm is that the representation learned by dimensionality reduction is embedded in the objective functions and will be simultaneously learned together with the clustering parameters.  Our framework is generic and could be adapted to different unsupervised learning tasks, without needing any data augmentation.

\section{Related Work}
\label{sec:intro}
%Unsupervised learning algorithms have become fundamental components in various machine learning and computer vision applications. Among these, clustering stands out as a crucial technique. Our work is closely related to two major aspects of unsupervised learning algorithm, i.e., clustering and dimensionality reduction.
 \subsection{Clustering}
Several traditional clustering algorithms are closely related to our work. k-Means, proposed by Lloyd \cite{lloyd1982least},  might be the most popular and widely used clustering algorithm. %It is a centroid-based clustering algorithm that aims to partition observations to $k$ clusters where each observation belongs to the nearest cluster. 
On the other hand, Expectation Maximization algorithm \cite{moon1996expectation} and its variants \cite{greff2017neural} are clustering and density estimation algorithms that are frequently applied to parameter estimation of mixture models. 
%The EM algorithm iteratively repeats two basic steps, namely the E-step and M-step, until convergence to find an optimal solution to the density estimation problem to minimize log-likelihood with latent variables. 
In this work, we consider using these two clustering algorithms to update some of the parameters and assign clustering labels.

\subsection{Dimensionality Reduction}
Dimensionality reduction has been a fundamental problem in machine learning and data analysis, with classical approaches like Principal Component Analysis (PCA) \cite{pearson1901liii} and Linear Discriminant Analysis (LDA) \cite{Tharwat2017LinearDA}  serving as baseline for modern techniques. The connection between dimensionality reduction and sparsity has significantly influenced various domains of machine learning. For instance, recent advancements in sparse tree generation \cite{dawer2020generating} demonstrate how principles from dimensionality reduction can be effectively adapted to hierarchical models. Similarly, neural network compression techniques \cite{guo2021study} can be interpreted as a learned form of dimensionality reduction, leveraging sparsity to reduce model complexity while preserving essential information.

% Several dimensionality reduction methods can be linked to the proposed model framework and algorithm. 
% Principal Component Analysis \cite{pearson1901liii} was firstly invented by Pearson in early 1900s and has long become the representative of linear dimensionality reduction methods. 
%It aims to find orthogonal linear projections called Principal Components for dimensionality reduction. 
Manifold learning has emerged as a crucial techniques for  dimensionality reduction. Maaten and Hinton \cite{van2008visualizing} proposed the T-distributed Stochastic Neighbor Embedding (T-SNE) that is particularly well-designed for high-dimensional datasets. T-SNE is now widely-used in numerous unsupervised learning tasks. Isomap \cite{balasubramanian2002isomap} and  Multidimensional Scaling (MDS) \cite{carroll1998multidimensional} are widely-used manifold learning algorithm aiming to preserve pairwise distances in low dimensional space.

In this work, T-SNE and PCA are considered as an initialization for the dimensional reduction in the proposed framework and algorithm.

\subsection{Deep Unsupervised Feature Learning}
Deep unsupervised learning of features have been a heated topic in recent research work related to computer vision and machine learning. 
Coates and Ng \cite{coates2012learning} applied k-means in pre-trained convnets and learned each layer sequentially.  
Caron  presented DeepCluster \cite{caron2018deep} that iteratively groups the features with K-means and then utilizes the subsequent assignments as pseudo-labels to update the weight of the network. In contrast, the clustering parameters and the network weights are updated simultaneously in our framework, by walking on the minimum clustering energy manifold.
Zhan \cite{zhan2020online} proposed Online Deep Clustering that maintains two dynamic memory modules, i.e. one to save labels and features, the other one to save centroid memory for centroid evolution.
 Chen  \cite{chen2020simple} introduced SimCLR as a simple framework for contrastive learning of visual representations. 
 Those algorithms mainly leverage complex Convolutional Neural Networks, such as VGG \cite{simonyan2014very}, AlexNet \cite{krizhevsky2012imagenet} and Resnet \cite{he2016deep} to acquire a representation for large-scale datasets.

 \subsection{Gaussian Mixture Model}
 Gaussian Mixture Models  (GMMs) \cite{reynolds2009gaussian} have been widely used in various clustering and density estimation tasks. They provide a flexible approach to modeling complex data distributions by combining multiple Gaussian components.  
Viroli \cite{viroli2019deep} proposed a deep GMM structure to learn complex relationship more efficiently. Liu \cite{liu2018unsupervised} proposed a novel clustering methods of combining GMM with uniform background noise. This method mainly focuses on the clustering accuracy on fore-ground positive classes other than background noise classes.

\section{Bi-level Optimization for Clustering}
Given a set of training examples $X = \{\bx_i\in \RR^p, i = 1,...,N\}$, our algorithm is aimed at obtaining an unsupervised version of the Linear Discriminant Analysis. For this goal, it has the following components: 

1. A dimension reduction function $A_\btheta$, which projects $X= \{\bx_i, i = 1,...,N\}\subset \RR^p$ to $X_\btheta = \{ \bx_i^\theta=A_\btheta(\bx_i), i = 1,...,N\}\subset \RR^d$. Typical choices of $A_\btheta$ include linear projection matrices, T-SNE, as well as more complex functions such as neural networks. The parameters set $\btheta\in \RR^l$ contains all the learnable parameters of $A$. For example, $\btheta$ could contain the weights and biases for all layers in a neural network. 

2. A clustering criterion $E(\bu, \btheta)=\sum_{i=1}^N e(\bu,A_\btheta(\bx_i))$ that is minimized with respect to the clustering parameters $\bu\in \RR^m$ to  cluster the projected points $X_\btheta$.
Through this minimization, the parameters $(\bu, \btheta)\in \RR^m\times \RR^l$ are restricted to lie on a manifold $M \subset \RR^{m+l}$.

\begin{equation}
M = \{(\bu_\btheta, \btheta)\in \RR^m\times \RR^l|  \bu_\btheta = \argmin_\bu E(\bu,\btheta),\forall \btheta\in \RR^l\}
\end{equation}

3. A clustering evaluation function $g(\bu)$ that needs to be optimized to obtain the best clustering parameters $\bu\in \RR^m$. The function $g(\bu)$ aims at separating the clusters as much as possible. For this purpose we could use a distance function between the corresponding densities parameterized by the cluster parameters $\bu$ e.g. the Kullback–Leibler Divergence or the Bhattacharyya Distance. 
See below for an example.

The goal of the optimization is to find the projection $\btheta$ and that best separates the clusters, i.e.:
\begin{equation}
(\hat \bu, \hat \btheta)=\argmax_{(\bu_\btheta,\btheta)\in M} g(\bu_\btheta) \label{eq:bilevel}
\end{equation}
This equation is called bi-level optimization because in involves two intertwined optimizations, the high level minimization of $g(\bu_\btheta)$ over $\btheta$, and the low level minimization over $\bu$ to obtain $\bu_\btheta = \argmin_\bu E(\bu,\btheta)$ for each $\btheta$.

For example, we can assume a Gaussian Mixture Model for the projected data $X_\btheta\subset \RR^d$, in which case:
\begin{equation} 
E(\bu, \btheta) =  -\sum_{i=1}^{n}\log \sum_{k=1}^{K}\pi_k \N( A_{\theta}(\bx_i); \bmu_k, \bSigma_k)
\label{eq:E}
\end{equation}
where $\N(\bz, \bmu,\bSigma)$ is a Gaussian with mean $\bmu \in \mathbb{R}^{d}$ and covariance $\bSigma$, 
\begin{equation}
\N(\bz; \bmu, \bSigma) = \det(\bSigma)^{-1/2}(2\pi)^{-d/2}\exp(-\frac{1}{2} (\bz - \bmu)^{T}\bSigma^{-1}(\bz-\bmu))
\end{equation}
and the clustering parameters are $
\bu = (\bu_1, ..., \bu_K)=(\bmu_1,\bSigma_1,\pi_1,...,\bmu_K, \bSigma_K, \pi_K)$.

 For $g(\bu)$, we adopt the Bhattacharyya distance, which measures the similarity between two distributions. The Bhattacharyya distance is related to Kullback–Leibler (KL) Divergence which is a well-known and widely used metric to measure to measure how one probability distribution is different form another one. Unlike KL Divergence, Bhattacharyya distance is symmetric and defined as:
\begin{equation}
BD(p,q) = -\ln \int \sqrt{p(X_\btheta)q(X_\btheta)} d\bx ,
\end{equation}
 where $p, q$ are two distinct probability density functions for $\bx_{\theta}$.
In the case of $K=2$, the Bhattacharyya distance between the two multivariate normal distributions $ N(\bmu_1, \bSigma_1)$ and $N(\bmu_1, \bSigma_1)$ has an analytic form: 

\begin{equation}
\label{eq:bd2}
\begin{split}
g(\bu) &= BD(\bu_1, \bu_2) \\
  &= \frac{1}{8}(\bmu_1-\bmu_2)^T\bSigma^{-1}(\bmu_1-\bmu_2)\\
  &+\frac{1}{2}\log(\frac{(\det(\bSigma))}{\sqrt{\det(\bSigma_1)\det(\bSigma_2)}})\\
\bSigma &= \frac{\bSigma_1+\bSigma_2}{2}
\end{split}
\end{equation}

\subsection{Gradient Based Manifold Optimization}

\begin{figure}
\vspace{-3mm}
\centering
\includegraphics[width=5cm]{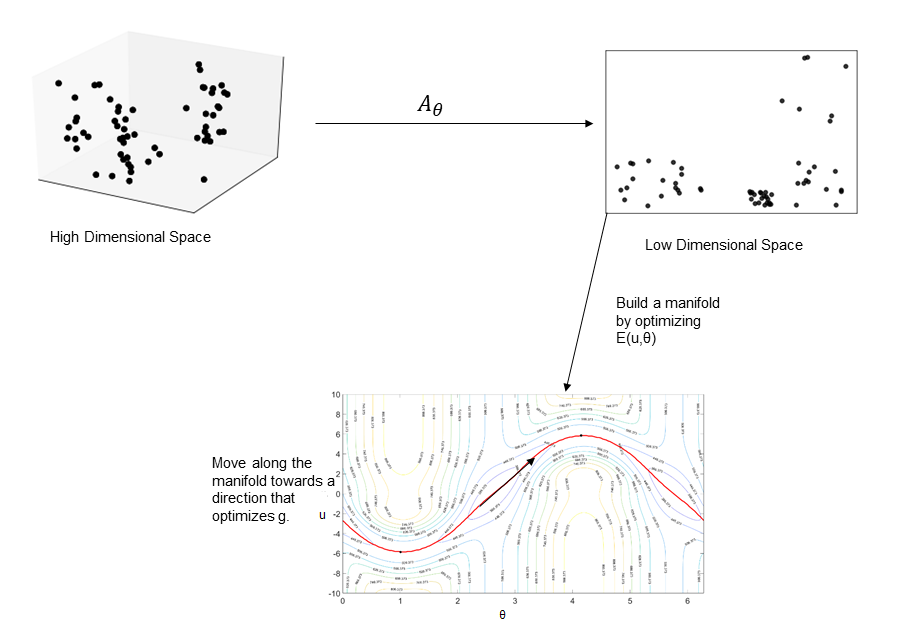}
\caption{Illustration of the gradient-based manifold optimization framework.
}\label{framework}
\vspace{-3mm}
\end{figure}

We adopt a gradient based approach to the optimization problem \eqref{eq:bilevel}. 
Figure \ref{framework} illustrates the proposed manifold optimization approach.

For that, we need the gradient of the high level function $g(\bu_\btheta)$ w.r.t. the low level parameters $\btheta$, which we obtain using the chain rule
\begin{equation}
\label{eq:chainrule}
\frac{\partial g}{\partial \btheta}=\frac{\partial g}{\partial \bu}\frac{\partial \bu}{\partial \btheta}=\nabla_\bu g \nabla_\btheta \bu_\btheta
\end{equation}

The gradient of $g(\bu)$ w.r.t. $\bu$ can be easily computed, especially when $g$ has an analytic form such as \eqref{eq:bd2}.

To compute the gradient of $\bu_\btheta$ w.r.t. $\btheta$, let $\bh =\nabla_\bu E(\bu,\btheta)$. Then we have
$\bh_j(\bu_\btheta,\btheta)=0$ for all $j$. Taking the derivative w.r.t. $\theta_i$ we get

\begin{equation}
\sum_k \frac{\partial \bh_j}{\partial u_k}\frac{\partial u_k}{\partial \theta_i}+\frac{\partial \bh_j}{\partial \theta_i} = 0
\end{equation}

which means
\begin{equation}
H(\bu_\btheta,\btheta) \frac{\partial \bu_\btheta}{\partial \theta_i}=-\frac{\partial^2 E(\bu,\btheta)}{\partial \bu \partial\theta_i}
\end{equation}

where $H(\bu,\btheta)$ is the Hessian of $E(\bu,\btheta)$ w.r.t. $\bu\in \RR^m$, i.e. $H_{ij}=\frac{\partial^2 E(\bu,\btheta)}{\partial u_i \partial u_j}$. Observe that this Hessian is $m\times m$, therefore it is not very large. For example for two 1D Gaussians, $\bu\in \RR^6$, so the Hessian is at most $6\times 6$.
We obtain:
\begin{eqnarray}
\nabla_\btheta \bu_\btheta = -H(\bu_\btheta,\btheta)^{-1}\frac{\partial^2 E(\bu,\btheta)}{\partial \bu \partial\theta}
\end{eqnarray}

Observe that $\nabla_\btheta \bu_\btheta$ and $\frac{\partial^2 E(\bu,\btheta)}{\partial \bu \partial\btheta}$ are both of size $m \times l$, where again $m$ is not very large, but $l$ could be large (e.g. could be millions for a neural network).

Using these gradients, we can start from an initial point on the manifold, and then follow the manifold by gradient update (with adaptive steps) until convergence or a maximum number of iterations has been reached.
We obtain a general Gradient Based Manifold Optimization Algorithm (GMOA), described in Algorithm \ref{alg:gmoa}.

\begin{algorithm}
	\caption{{\bf  Gradient Based Manifold Optimization Algorithm (GMOA)}}
	\label{alg:gmoa}
       {\bf Inputs}: Training set $X = \{\bx_i\}_{i = 1}^{N}$, initial parameters $\btheta_0$, learning rate $\eta_0$, number of iterations $N^{iter}$.

{\bf Outputs}: Trained projection function $A_{\btheta}$, high level parameters $\bu$.

\begin{algorithmic} [1]
\STATE Update $\bu_{0}$ by minimization: $\bu_{0} =  \argmin_{\bu} E(\bu,\btheta_{0})$
\FOR {$t=0$ to $N^{iter}$}
           \STATE Set $\eta = \eta_0$
		\STATE Compute $g_{t} = g(\bu_{t})$  
		\STATE Compute $\frac{\partial^2 E}{\partial \bu \partial\btheta}(\bu_t,\btheta_t)$, Hessian $H(\bu_t,\btheta_t)$, and  gradient $\nabla_\bu g(\bu_t)$
		\STATE Compute
\begin{equation}
\Delta_t=-	\nabla_\bu g(\bu_t) H(\bu_t,\btheta_t)^{-1}\frac{\partial^2 E}{\partial \bu \partial\btheta}(\bu_t,\btheta_t)\label{eq:gdmoa}
\end{equation}		
		  %based on \ref{eqn: gtomu}, \ref{eqn: gtosigma}, \ref{eqn: gtopi} and \ref{eqn:secondorder} 
		\STATE Update $\btheta_{t+1} = \btheta_{t} + \eta \Delta_t, \bu_{t+0.5}=\bu_t+\eta \nabla_\bu g(\bu_t)$
		\STATE Update $\bu_{t+1}$ by minimization: $\bu_{t+1} =  \argmin_{\bu} E(\bu,\btheta_{t+1})$ with warm start $\bu_{t+0.5}$
		\STATE Compute $g_{t+1} = g(\bu_{t+1})$. 
%		\IF {$g_{t+1} < g_{t} - \epsilon$}
%		\STATE Shrink the learning rate $\eta = \eta/2$ and go back to Step 7
%		\ENDIF  
\ENDFOR
	\end{algorithmic}
\end{algorithm}

Per Algorithm \ref{alg:gmoa}, $\btheta$ is initialized with $\btheta_0$ and $\bu$ is initialized by finding the manifold location for $\btheta_0$ by energy minimization. After this initialization, the paired parameters $(\bu_0, \btheta_0)$ will be on the manifold. Then, $\btheta$ will be updated in the direction that maximizes $g$ by gradient ascent. Since the update of $\btheta$ might push $(\bu, \btheta)$ out of the manifold, Step 8 of the algorithm intends to drive the paired parameters back to the manifold.

We consider the specific details for applying Algorithm \ref{alg:gmoa} to jointly learn a projection function and clustering parameters for Gaussian Mixture Models.
Gaussian Mixture Models are extensively used in density estimation and clustering.  
The clustering parameters in this case are 
\[
\bu = (\bu_1, ..., \bu_K)=(\bmu_1,\bSigma_1,\pi_1,...,\bmu_K, \bSigma_K, \pi_K) 
\]
\[
\text{ so } \bu_i=(\bmu_i,\bSigma_i,\pi_i), i=1,...,K.
\]
More details are specified in next subsection.
\subsection{Manifold Optimization for Gaussian Mixture Models}
In the last sub-section, we discussed a generic gradient based manifold optimization algorithm. 
In this section, we will consider the specific details for applying GMOA to jointly learn a projection function and clustering parameters for Gaussian Mixture Models.

Gaussian Mixture Models are extensively used in density estimation and clustering.  
The clustering parameters in this case are 
\[
\bu = (\bu_1, ..., \bu_K)=(\bmu_1,\bSigma_1,\pi_1,...,\bmu_K, \bSigma_K, \pi_K), \]
\[
\text{ so } \bu_i=(\bmu_i,\bSigma_i,\pi_i), i=1,...,K.
\]

For an observation $\bz \in \mathbb{R}^{d}$, the probability density function (pdf) for a Gaussian Mixture Model is given by:
\begin{equation} 
\label{equ: gmm}
f(\bz)=\sum_{k=1}^{K}\pi_k \N(\bz; \bmu_k, \bSigma_k),
\end{equation}
obtaining the negative log likelihood  as $E(\bu,\btheta)$.

For $g(\bu)$ we use the mean of the Bhattacharyya distances between all pairs of components $\bu_i,\bu_j,i<j$, to encourage all Gaussians to be far from each other. There is one more issue with GMMs, namely that when a mixture component becomes degenerate (has less than $d$ elements), the negative log likelihood $E(\bu,\btheta)$ becomes infinite. We discourage that by discouraging each $\pi_k$ to be close to 0. For these purposes we use the following
\begin{equation} 
\label{eq:lossgmm}
g(\bu)=\frac{2}{K(K-1)}\sum_{i=1}^K\sum_{j=i+1}^K BD(\bu_i,\bu_j)+\sum_{i=1}^K \log \pi_k
\end{equation}

The GMMs have also extra constraints that each mixture weight $\pi_k$ is positive and $\sum_i=1^K \pi_k=1$. Thus our bi-level optimization problem for GMM is:
\begin{equation} 
\begin{split}
         & (\hat \bu, \hat \btheta)=\argmax_{(\bu_\btheta,\btheta)\in M} g(\bu) \\
           \bu_\btheta &= \argmin_{\bu} -\sum_{i=1}^{n}\log \sum_{k=1}^{K}\pi_k N(\bmu_k, \Sigma_k,  A_{\btheta}(\bx_i))\\
           & \text{s.t.}\qquad \sum_{k}{\pi_k} =1\\
           & \text{s.t.}\qquad \pi_k \ge 0 ,\text{ for all k}\label{eq:gmmbilevel}
\end{split}
\end{equation}

We can still use GMOA for solving \ref{eq:gmmbilevel}, with the modification that the minimizations in Steps 1 and 8 are constrained minimizations.

Currently, we consider each Gaussian in GMM as isotropic, therefore

\begin{equation}
\bSigma_k = \diag(\bsigma_{k}) = \begin{bmatrix}
 \sigma_{k1}^2, \cdots, 0 \\
\cdots, \cdots, \cdots\\
 0 ,\cdots, \sigma_{kd}^2 
\end{bmatrix}
\end{equation}
where each $\bsigma_i\in \RR^d$ is a vector. Then the clustering parameters are

\begin{gather} 
\label{eqn: final u}
\bu = (\bmu_1, \bsigma_{1},\pi_1, \bmu_2, \bsigma_{2}, \pi_2, ..., \bmu_K, \bsigma_{K},\pi_K)
\end{gather}

The gradients of $g(\bu)$ are:

 \begin{equation}
\label{eqn: gtomu}
 \frac{\partial g}{\partial \bmu_k} = -\frac{2}{K(K-1)} \sum_{i=1, i\not =k}^{K} (\frac{\bSigma_i + \bSigma_k}{2})^{-1}(\bmu_k - \bmu_i)
\end{equation}

 \begin{equation}
\label{eqn: gtosigma}
 \frac{\partial g}{\partial \bsigma_{k}^2} = \sum_{i=1, i \neq k}^{K} \frac{1}{8} \left(\frac{(\bmu_{k} - \bmu_{i})^2}{(\bmu_{k}+\bmu_{i})^2}-\frac{2}{\bmu_{k}+\bmu_{i}}-\frac{1}{\bmu_{k}} \right),
\end{equation}

 \begin{equation}
\label{eqn: gtopi}
 \frac{\partial g}{\partial \pi_k} = -1/\pi_k
\end{equation}
where $k \in \{ 1,\cdots, K \}$.

For the calculation of  $\frac{\partial^2 E(\bu,\btheta)}{\partial \bu \partial\btheta}$ and $\bh(\bu,\btheta) =\nabla_\bu E(\bu,\btheta)$, the analytic solutions are difficult to obtain. 

We therefore used central difference approximations for the first order derivatives to approximately calculate  $\bh(\bu,\btheta)$,
 \begin{equation}
\label{eqn:firstorder}
 \bh(\bu,\btheta)   \approx \frac{E(\bu+\Delta \bu, \btheta)- E(\bu-\Delta \bu,\btheta)}{2||\Delta \bu||}
\end{equation}
Specifically, the $j$-th component of $ \bh(\bu,\btheta)$ is approximated as
 \begin{equation} \label{eq:hj}
 h_j(\bu,\btheta)  \approx \frac{E(\bu+\delta \be_j, \btheta)- E(\bu-\delta \be_j,\btheta)}{2\delta}
\end{equation}
where $\delta$ is a small number close to 0 (e.g. $\delta=0.00001$) and $\be_j = (0,0,\cdots,1,\cdots,0,0)$, with the $j$th element being $1$, and the rest zero.

We then further calculated 
\begin{equation}
\frac{\partial^2 E(\bu,\btheta)}{\partial u_j \partial\btheta}  =  \frac{ \partial h_j(\bu,\btheta)}{\partial\btheta}, \forall j \in \{1,\cdots, m\}
\end{equation}
by back-propagation on the
 $h_j(\bu,\btheta)$ approximation \eqref{eq:hj} with respect to $\btheta$, obtained by using Pytorch's automatic differentiation capability. 

For estimation of  Hessian matrix $H(\bu,\btheta)$, a Pytorch Toolbox \footnote{Link: https://github.com/mariogeiger/hessian} has been used.

In each iteration, once $(\btheta, \bu)$ have been updated in Step 7 of Algorithm \ref{alg:gmoa}, $\bu$ will be further updated by EM per Step 8 to guarantee that the solution trajectory is moving along the manifold. For a faster convergence, after a few EM iterations, the LBFGS algorithm can be used to find the optimal GMM solution faster.
\section{Implementation Details}
In this part, we will discuss implementation details for our proposed Algorithm \ref{alg:gmoa} for the GMM settings.

At first, we leverage some initialization methods to initialize $\bu$ and $\btheta$. 
For $\btheta$, when $A_{\btheta}$ is a linear projection matrix from $\mathbb{R}^{p}$ to $\mathbb{R}^{d}$, $\btheta$ can be initialized using linear dimension reduction methods such as PCA with principal components as projection vectors. 
If $A_\btheta$ is a feed-forward Neural Network (NN), we will apply a non-linear dimension reduction technique such as T-SNE as initialization. 
Then, the NN is pre-trained by minimizing the Mean Square Loss between NN output and TSNE projection output. 
By these means, the NN is initialized to mimic the effect and consequences of T-SNE in terms of dimension reduction. 

As for the NN proposed in the model, we try some simple Neural Network structures of FNN with one to three hidden layers. 
In the experiments discussed in the following sections, a typical NN utilizes $$\{ \text{dimension of input}, 64, 32, \text{dimension of projected space} \}$$ as numbers of neurons in each  layers. While updating the $A_{\btheta}$ in Algorithm \ref{alg:gmoa}, the parameters in the last fully-connected layer of the FNN is adjusted to normalize the final output with mean $0$ and standard deviation $1$ as follows:

\begin{equation}
\label{eq:norm1}
\hat{\bb} =\frac{\bb -\bmu_L}{\bsigma_L}, \;
\hat{\bbeta} = \frac{\bbeta}{\bsigma_L}, 
\end{equation}
where $\bmu_L$ is the mean of the final projection obtained by the Neural Network and $\bsigma_L$ is the standard deviation of final projection prior to normalization. $\bb$ and $\bbeta$ in \ref{eq:norm1} are the bias and weight parameters in the last fully-connected layer of the NN. 

In terms of training process,  an adaptive learning rate $\eta$ is applied if necessary. 
 $\bu$ is further updated and adjusted to ensure the parameter pair $(\bu,\theta)$ is on the manifold. 
%To check this,  we compare $g(\bu)$ before and after the updates with a pre-defined acceptance threshold $\epsilon$ as reflected in Step 9 to Step 11 in Algorithm \ref{alg:gmoa}. 
%By this means, the learning rate $\eta$ becomes adaptive to keep a balance between training speed and appropriate updating trajectory. 
In the real data experiments, we generally applied an initial learning rate for $\eta= 0.005$. 

We generally applied EM algorithm with a warm start in Step 8 of proposed Algorithm. 
 In various experiments, we tried 50, 100, or 200 iteration steps in EM applied in proposed Algorithm . 
To accelerate the convergence of EM, we applied L-BFGS \cite{liu1989limited} with the outcome of several EM steps as initialization.

Ultimately, by learning and updating $\theta$ in proposed algorithm , the dimension reduction is accomplished. 
The parameter vector $\bu$ represents estimations of important statistics (i.e. mean, variance and weight) for each cluster. 
Hence, a natural way to cluster the projected data based on $\bu$ under the GMM framework is to calculate the log- likelihood of Gaussian of each cluster with a data point $\bx_i$ and assign it the label where the log-likelihood is maximized. This clustering algorithm is defined below:

\begin{algorithm}[htb]
	\caption{{\bf Clustering Based on Output of GMOA}}
	\label{alg:clusteing}
       {\bf Inputs}: Training Set $ X = \{\bx_i\}_{i = 1}^{N}$, $\bu$ and $\btheta$ learned from GMOA

{\bf Outputs}: Clustering labels  $\{y_i\}_{i = 1}^{N}$
	\begin{algorithmic} [1]

		\FOR {$i = 1$ to $N$ }
		\FOR { $k= 1$ to $K$}
		\STATE Calculate   $f(i,k) = N(\bmu_k,\Sigma_k, A_{\btheta}(\bx_i))$
		\ENDFOR
		\STATE Obtain label $y_i = \argmax_{k \in \{1, \cdots, K\}} \{f(i,k)\}_{k=1}^{K}$
		\ENDFOR
	\end{algorithmic}
\end{algorithm}

We also applied various clustering algorithms (i.e. K-means, EM) directly on the projected inputs (i.e. $A_\btheta(\bx)$ learned from GMOA). 

\section{Experiments}
In this section, we perform numerical experiments on synthetic GMM datasets as well as the well-known MNIST public dataset \cite{lecun1998mnist}).  We utilized the Pytorch \cite{paszke2019pytorch} framework to train the GMOA algorithm.

\subsection{2D GMM Simulated Data}
In this section, the test results are presented on GMM simulated in 2D ($p=2$) for a good visualization of the algorithm. 
The projected space dimension is $d=1$. 
The dimension reduction is performed by a linear projection vector in $\mathbb{R}^{p\times 1}$.

For $2D$ GMM data, we consider a very simple case. 
Suppose a GMM with two Gaussian components are generated in 2D and they are projected to $1D$ with a projection matrix $A = [a_1, a_2]$. Without loss of generality, $A$ can be formulated as $A = [\cos(\theta), \sin(\theta)]$. 
we have:

\begin{gather} 
\label{eq:u2d}
\bu = (\bmu_1, \bsigma_1,\pi, \bmu_2, \bsigma_2, 1- \pi)
\end{gather}

For better visualization, we consider the special case when the GMM cluster variances and cluster weights are known. 
Under such circumstances, the parameters to be estimated are $\bmu_1,\bmu_2$ and $\btheta$. 
Based on \ref{eq:bd2}, to optimize $g$, it is equivalent to optimize $|\bmu_1-\bmu_2|$.  Let $\Delta_\bmu=\bmu_1-\bmu_2$.

The dataset is composed of a mixture of Gaussian with two clusters in 2d. The true parameters for the mixtures are $\bmu_1^{*} =[0,0]$, $\bmu_2^{*} = [-3,-5]$, $\Sigma_1 = \begin{bmatrix} 
2 & 0 \\
0 & 2 
\end{bmatrix}$, $\Sigma_2= \begin{bmatrix} 
1 & 0 \\
0 & 1 
\end{bmatrix}$, $\pi_1 = \pi_2 =0.5$. 

Therefore, our task is to address the following optimization problem:
  \begin{align}
         & \max  |\bmu_1 - \bmu_2|\\
           (\bu,\btheta) &= \argmin_\bu E(\bu,\btheta)
\end{align}

\begin{figure}[htbp]
\centering
\includegraphics[height=4.5cm]{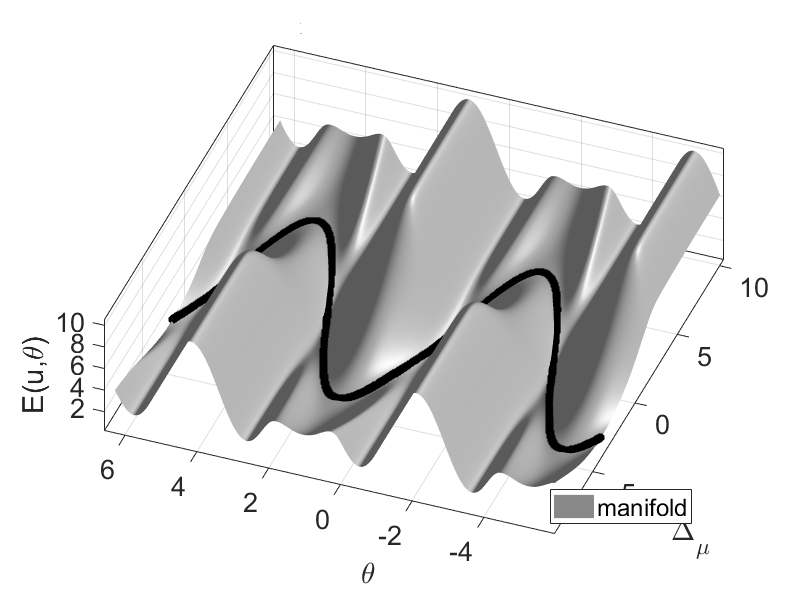}
\vskip -3mm
\caption{Surface and manifold plot of $(\Delta_\bmu, \theta, E(\bu,\btheta))$ for 2D Gaussians.}\label{2d_GMM_surface}
\vspace{-3mm}
\end{figure}
We compute $E(\bu, \btheta)$ on a grid of values $(\Delta_\mu, \btheta)\in [-10,10]\times [2\pi,2\pi]$. 
The surface of $E(\bu, \btheta)$ and the manifold $M$ are are shown in Figure \ref{2d_GMM_surface}.

From the plot, for $\Delta_\bmu \in [-10,10]$ and $\btheta \in [-2\pi, 2\pi]$, the manifold is the black solid line on the surface. The optimal solution points are the points on the black line where $|\Delta_\bmu|$ is maximized. From this data, we obtain four optimal solutions, 

\begin{figure}[ht]
\vspace{-3mm}
\centering
\includegraphics[height =3cm]{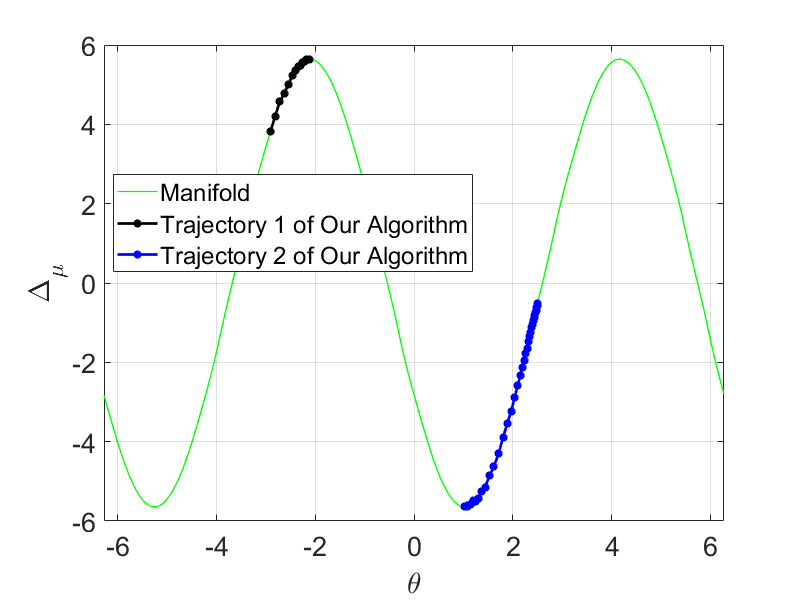}
\vskip -3mm
\caption{Trajectory of our algorithm in 2D.}\label{2d_GMM_trajectory}
\vspace{-2mm}
\end{figure}
Our algorithm will get to one of the optimal solutions, as illustrated by the trajectory plots from Figure \ref{2d_GMM_trajectory}. 
There we started from $\btheta = 2.5$ and $\btheta = -3$, and our algorithm finally reached one of the optimal solutions $(5.64,-2.12)$ and $(-5.64, 1.02)$ for $\Delta_\bmu, \btheta$. 
As shown in Figure \ref{2d_GMM_trajectory}, the trajectory of our algorithm is in line with our expectation as it moves along the manifold.

\begin{figure}[htbp]
\vspace{-3mm}
\centering
\includegraphics[width = 4.5cm]{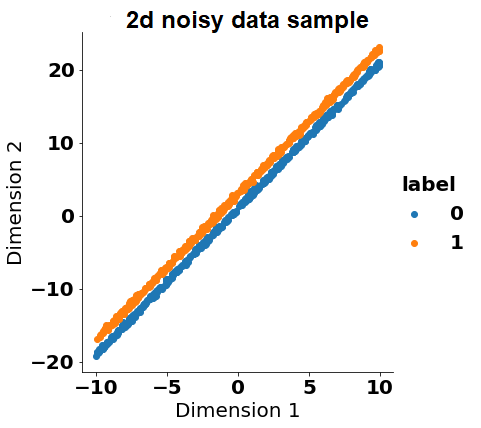}
\caption{Example of data with noise in 2D.}\label{2d_noisy_GMM}
\vspace{-2mm}
\end{figure}

To test the robustness of algorithm on a noisy dataset, we adopt a 2d noisy data and compared the performance of our proposed algorithm with other regular clustering algorithms. 
We generate the 2D noisy data in the following way: first, we generate two parallel lines in 2D with same slope and different intercepts. 
Then, we sample points from the two lines and add random Gaussian ($N(0,\epsilon)$) noise to the second dimension of each point. The data generated is plotted in Figure \ref{2d_noisy_GMM}.

The original data in 2d may be separable by applying PCA to $1D$. To fairly compare our algorithm with other popular clustering algorithm, we use a random initialization to project the data in 2D to $1D$. 

The clustering algorithms being compared are: K-means \cite{arthur2007k}, EM \cite{moon1996expectation}, Agglomerative Clustering \cite{carlsson2010characterization} and Spectral Clustering (\cite{shi2000normalized, ng2002spectral}).
For a fair comparison, we tuned the following parameters in the clustering algorithms that were compared: for K-means, we used K-means ++ as initialization method. For Spectral Clustering, the number of neighbors was chosen from $\{1,5,10,20\}$ and for affinity matrix, we used RBF and Nearest Neighbors. 
For Agglomerative Clustering, we tried several linkage methods, i.e. Ward, Completed, Single Linkage. 
As for EM, 10 random initializations were used and the best result was considered. 
\begin{figure}[htb]
\vspace{-3mm}
\centering
\includegraphics[height=2cm]{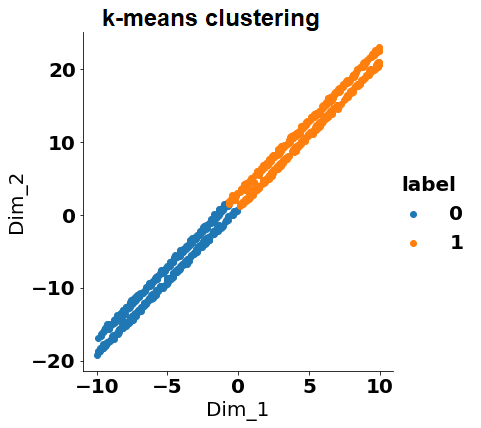}
\includegraphics[height=2cm]{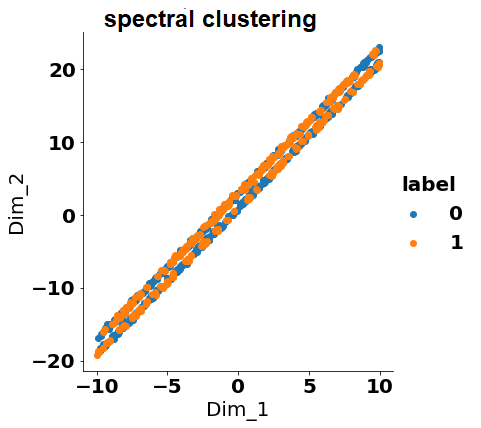}
\includegraphics[height=2cm]{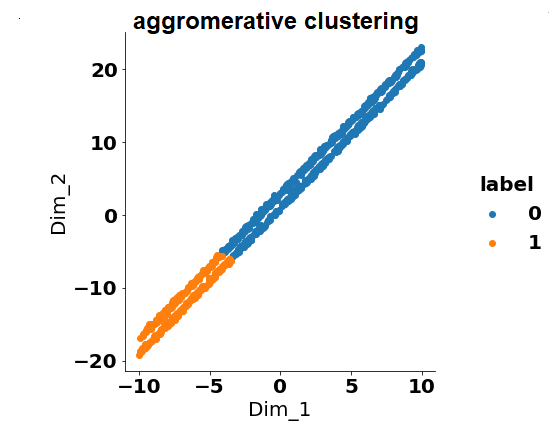}
\includegraphics[height=2cm]{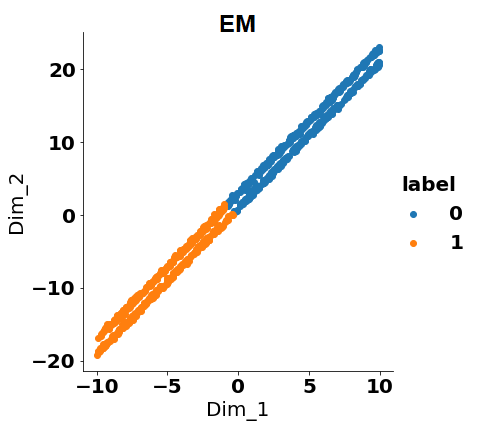}
\includegraphics[height=2cm]{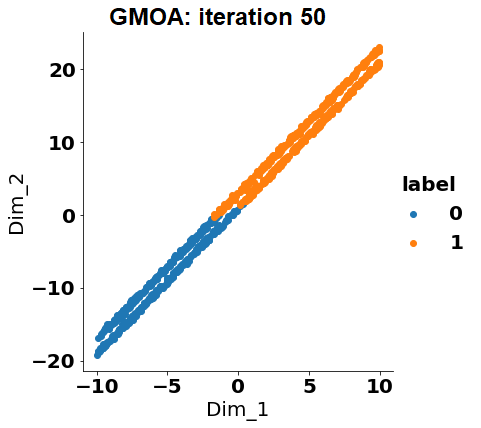}
\includegraphics[height=2cm]{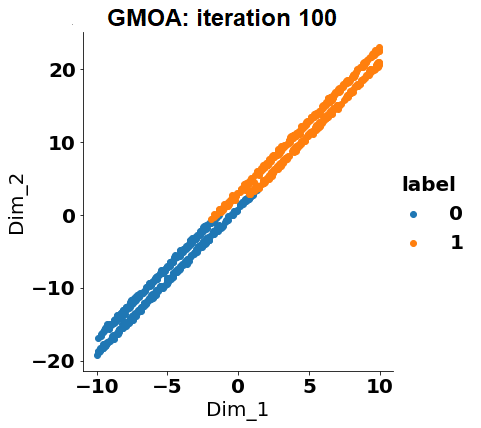}
\includegraphics[height=2cm]{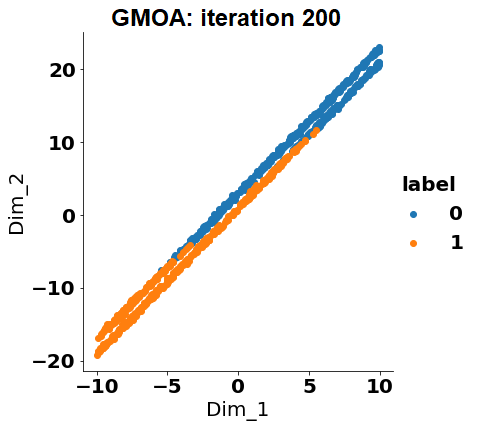}
\includegraphics[height=2cm]{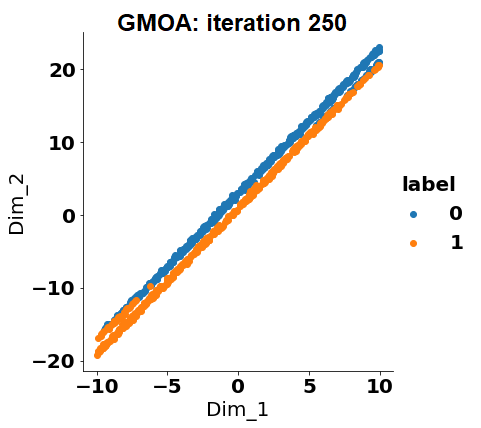}
\includegraphics[height=2cm]{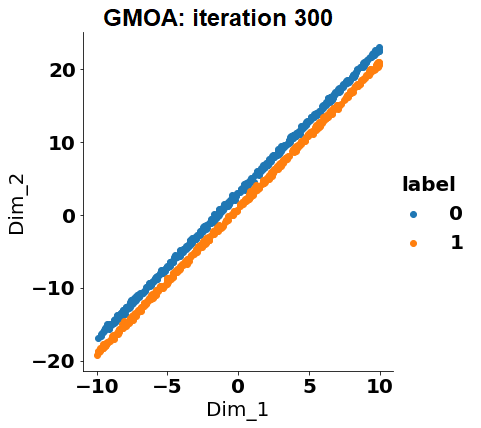}
\vspace{-3mm}
\caption{Comparison of the proposed algorithm with other popular clustering algorithms on 2D noisy data.}\label{fig:noisy_2d}
\vspace{-3mm}
\end{figure}

The obtained results are shown in Figure \ref{fig:noisy_2d}. 
Based on the clustering results, it takes around 300 iterations for the proposed algorithm to obtain a promising result from a random initialization. In summary, our algorithm is somehow robust to noise as shown in this experiment.
\subsection{3D GMM Simulated Data}

In this section, we simulated a $p=3$ dimensional GMM dataset to illustrate our algorithm's behavior. 
The projected space is in $d=1$. 
The data simulation is similar to 2D GMM data simulation specified in the last subsection. The mean vector for the 3D GMM is 
$\pi_1 = \pi_2 =0.5$, 
  $$\bmu_1 =[0,0, 0], \; \bmu_2 = [-3,-5,10]$$
  $$ \Sigma_1 = \begin{bmatrix} 
2 & 0 & 0 \\
0 & 2 & 0 \\
0& 0 & 2
\end{bmatrix}, \Sigma_2= \begin{bmatrix} 
1 & 0 &0 \\
0 & 1 &0 \\
0& 0& 1
\end{bmatrix}.$$

Like the simplest case in 2D GMM, we consider a special case in 3D that all covariance matrices and weight vectors of the GMM are known so that one can visualize the trajectory of our algorithm. 
Without loss of generality, $A$ can be formulated as 
$A = (\cos(\theta_1), \sin(\theta_1)\cos(\theta_2), \sin(\theta_1)\sin(\theta_2))$.

 Under such circumstances, the parameters to be estimated are $\bmu_1,\bmu_2$ and $\btheta=(\theta_1, \theta_2)$.Tto optimize $g$, it is equivalent to optimize $|\bmu_1-\bmu_2|$. Hence,  we denote $\Delta_{\bmu} = \bmu_1 - \bmu_2$.  

\begin{figure}
\centering
\includegraphics[width=5cm]{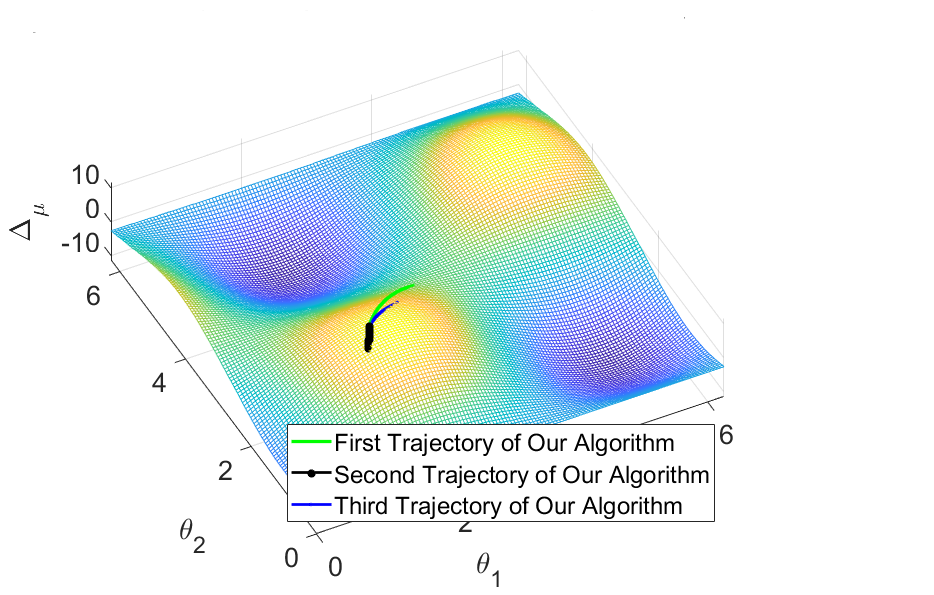}
\vskip -3mm
\caption{Surface and manifold plot of $(\Delta_\bmu, \btheta, E(\bu,\btheta))$ of 3D Gaussian (trajectory plots from three different $\btheta$  initializations {(3.0,3.0), (2.5,2.5), (1.6,1,6))}}
\label{3d_GMM_surface}
\end{figure}

We calculate and plot $E(\bu, \btheta)$ based on a grid of $\Delta_{\bmu}$ and $\btheta$ values and found the manifold point $\Delta_\mu$ for each $\btheta$. 
In Figure  \ref{3d_GMM_surface} is shown the obtained manifold and trajectories of our our algorithm from several random initializations of $\btheta$. 

As can be observed from Figure \ref{3d_GMM_surface}, for three experiments with various initialization, the final results of three generated paths converge to around $(1.8, 2.0, 11.5)$ for $(\theta_1, \theta_2, \Delta_{\bmu})$, which is an optimal solution to our optimization. 
The third initialization $(1.6, 1.6)$ for $\theta$ is much closer to the optimal solution for $\btheta_1, \btheta_2$, so it  converges quickly to optimal solution in just several iterations. 
For the other two initializations, it takes more iterations to converge but the algorithm converges fast in general, with fewer than 20 iterations to converge for this dataset.

In summary, from the evaluation on 2D and 3D GMM simulated data, the model performance is in line with our expectation in general. 
We constructed the manifold by fixing $\btheta$ and selecting $\bu$ that minimize $E(\bu,\btheta)$. 
The parameters to be estimated were simplified to $\Delta_\mu$ and $\btheta$. 
The trajectory of $(\Delta_\bmu, \btheta)$ of our algorithm was plotted based on updates of each iteration step. 
The trajectories plotted demonstrate that our algorithm can move along the manifold and eventually reach one of the optimal solution points in the manifold that optimize the objective function.

\subsection{Real Data}

In this subsection, we apply Algorithm \ref{alg:gmoa} to the  Modified National Institute of Standards and Technology (MNIST) data set. 
The MNIST dataset is a large database of handwritten digits (from 0 to 9) provided by Lecun and Cortes \cite{lecun1998mnist}. 
It contains 50K training samples, 10K validation samples and 10K testing samples. 
Each data sample is a black-and-white image of size $28\times28$ pixels. 

We applied our algorithm to the MNIST training dataset. 
First, we vectorized the input data to $784\times1$ dimension. 
Then, we used a two-layer Neural Network as $A(\btheta)$, with 64 neurons in the hidden layer, to project the input $\bx$ to a low-dimensional space, either 2D or 10D.  
Tests are performed on three digits (3, 5 and 9) of the MNIST data and on all ten digits respectively. 

The accuracy is computed by finding the best assignment of clusters to labels using the Hungarian algorithm \cite{kuhn1955hungarian} and then computing the accuracy of the obtained labeling.

For initialization, we employed T-SNE \cite{van2008visualizing} to obtain an initial clustering of the projected data $\bx_p$. 
After that, the parameters $\theta$ of  $A_\btheta$ (which is the two layer NN described above) are initialized by minimizing the Mean Square Loss between the T-SNE output and $\bx_p$. Finally, our algorithm is run on the MNIST dataset with different numbers of digits based on the mentioned initialization.

\begin{table}[t]
\center{
\scalebox{0.68}{
\begin{tabular}{|l|c|c|c|c|c|}
\hline
&K-means &EM &Spectral Clust. &DBSCAN &Our (GMOA)\\
\hline
\multicolumn{4}{l}{3 digits (3, 5, 9)}\\
\hline
Training Accuracy (\%)    &93.55 (29.38)  &88.44 (46.17) &97.15 &95.04 & {\bf 97.42 }\\ 
Test Accuracy (\%) &90.45 (30.64)   &88.33 (31.40)   &92.23 &89.21 & {\bf 96.43 }\\ 
\hline
\multicolumn{4}{l}{10 digits} \\
\hline
Training Accuracy (\%)  &83.33 (54.32) &75.39 (38,77) &81.99 &85.30  &{\bf 91.45}\\ 
Test Accuracy (\%) &71.72 (53.37)  &70.00 (42.69)  &74.11  &78.77 &{\bf 87.83}  \\
\hline
\end{tabular}}}

\caption{Accuracy of clustering algorithms  on the original MNIST dataset (projection to 2D. The results in bracket are the results on original $784D$ which are not applicable to Spectral Clustering and DBSCAN due to limit of memory).}\label{MNIST_alg2}
\end{table}
We compared our algorithm with several classic clustering algorithms, such as K-means \cite{arthur2007k}, EM \cite{dempster1977maximum}, Spectral Clustering \cite{vempala2004spectral} and DBSCAN \cite{ester1996density}. 
 These methods were applied on the same projected data obtained by T-SNE and on the original $784D$ data if memory allows. 
 Our algorithm was run for 50 epochs. For each algorithm are reported the best training accuracy and best test accuracy in Table \ref{MNIST_alg2}. 
For each clustering algorithm, we assume that the true number of clusters $k$ is known. 
The results in Table \ref{MNIST_alg2} are the test results on 3 digits and all 10 digit original MNIST data on a 2D projected space. The results in parentheses are the clustering results on the original 784D data.

From the comparison results, our algorithm outperforms the other popular clustering algorithms in terms of both training accuracy and test accuracy.

Aside from using T-SNE as initialization, we also conduct experiments by applying our method on a state-of-art visual representations based on a state-of-art method called SimCLR \cite{chen2020simple}. 
SimCLR is an unsupervised method that learns a semi-invariant representation for images such that the feature vectors extracted from the same image when perturbed by certain operations such as crop and blur are more similar than feature vectors obtained from different images under same perturbations. In this work, we used a CNN with two Convolutional layers (with $9\times 9$ and respectively $7\times 7$ filters) followed by $2\times 2$ max pooling stride 1, followed by one $7\times 7$ convolutional CSN Layer \cite{barbu2021compact} followed by $6\times 6$ max pooling as the encoder of the SimCLR.

Table \ref{MNIST_repre} presents the learned results with initialization on MNIST in 256D based on SimCLR followed by a projection from 256D to 10D using UMAP \cite{mcinnes2018umap}. The results for directly clustering in 256D are shown in parentheses.
Our algorithm GMOA is successful in further improving  the clustering results of the $SimCLR + UMAP$ representation on the training set as well as the test set. 

The hyper-parameters are initialized and tuned in the following ways: 1) $\theta_0$: We have mentioned that it is determined by T-SNE intialization. 2) learning rate and number of iterations: We tried different combinations. A general combination used in the experiments is to run 200 epochs with learning rate $10^{-4}$.

\begin{table}[htbp]
\center{
\scalebox{0.72}{
\begin{tabular}{|c|c|c|c|c|c|}
\hline
 &K-means &EM &Spect. Clust. &DBSCAN &Our (GMOA)\\
\hline
\multicolumn{4}{l}{3 digits (3, 5, 9)}\\
\hline
Training Accuracy (\%)    &97.58 (75.35) &97.45 (62.40) &98.88& 97.88& {\bf \bf 99.10}\\ 
Test Accuracy (\%) &96.63 (80.63)  &96.55 (83.62)   &97.62 &97.56& {\bf 98.69 }\\ 
\hline
\multicolumn{4}{l}{10 digits} \\
\hline
Training Accuracy (\%)  &96.25 (78.51)  &96.30 (62.65) &96.38 &97.18  &{\bf 98.24}\\ 
Test Accuracy (\%) &96.00 (83.64)  &95.99 (86.13)  &88.98  &96.10 &{\bf 96.91}  \\
\hline
\end{tabular}}}
\caption{Accuracy of clustering algorithms on the  MNIST dataset using the SimCLR representation (projection to 10D, the results in bracket are the results on 256D which are not applicable to Spectral Clustering and DBSCAN due to memory limitations).}\label{MNIST_repre}
\end{table}
\vspace{-3mm}

We also compared the performance of our algorithm with some other state-of-art deep clustering algorithms on entire MNIST datasets measured by clustering accuracy on the test set. The algorithms considered are GAN \cite{mirza2014conditional}, DEC \cite{xie2016unsupervised}, Deep Cluster \cite{caron2018deep} and DynAE \cite{mrabah2020deep}. 
Those methods usually utilize relevant deep convolutional neural network to generate meaningful representations.
The purpose of the test is to compare the performance of our algorithm with state-of-art clustering algorithms with constraints on the complexity of the underlying network structure. To achieve this goal, we apply the above mentioned algorithms on simple neural networks with one hidden layer for the fully-connected networks and up to two convolutional layers for convolutional networks. 

\begin{table}[ht]
\center{
\scalebox{0.72}{
\begin{tabular}{|l|c|c|c|c|c|}
\hline
  &GAN &DEC  &Deep Cluster &DynAE &Our (GMOA)\\
\hline
\multicolumn{4}{l}{10 digits}\\
\hline
Test Accuracy (\%) &82.8  &84.3   &65.6  &84.5& {\bf 87.9 }\\ 
\hline
\end{tabular}}}
\caption{Test accuracy of clustering algorithms on the  MNIST dataset using a shallow neural network structure.}\label{MNIST_comp}
\vspace{-3mm}
\end{table}
%------------------------------------------------------------------------- 
Table \ref{MNIST_comp} presents the learned results on MNIST for those approaches. The test results show that our algorithm (GMOA) outperforms other state-of-art deep clustering algorithms given a shallow neural network structure. Our algorithm does not rely on such a deep network structure compared to those algorithms. The application of the proposed framework on a deep Convnet will be the a potential future work for us.

\section{Conclusion}
%In this paper, we presented a novel framework for jointly learning Dimension Reduction and Clustering using a straightforward gradient optimization method. 
%The paper also presented a concrete application of the proposed framework by utilizing Gaussian Mixture Models, Bhattacharyya distances and linear projections or neural networks. 
%The ideal trajectory of our algorithm is to move along the desired manifold based on the maximization of the GMM log-likelihood, to reach an optimal point on the manifold that maximizes the inter-cluster distance function. 
%We performed various experiments on synthetic and real datasets and showed that our algorithm is useful for finding proper cluster assignments together with an unsupervised representation learned simultaneously by the proposed bi-level optimization algorithm. The performance of the algorithm depends on a good initialization, which we achieved using T-SNE or SimCLR+UMAP.

%We also observed some limitations in our current algorithm. Currently we have not employed very deep network structure in the dimension reduction function component. Thus, we compared our performance of our algorithm with some state-of-art algorithms with relatively shallow network settings. As we mentioned, current proposed algorithm is an application for the bi-level optimization framework. The experiments with setting of shallow network shows that it is promising to extend the current methodology to deeper neural network which could be future work for this study.

In this paper, we presented a novel framework for jointly learning dimensionality reduction and clustering using a gradient-based optimization method. We also demonstrated a concrete application of the proposed framework using Gaussian Mixture Models, Bhattacharyya distance, and both linear projections and neural networks. The algorithm follows a trajectory along a learned manifold by maximizing the GMM log-likelihood, aiming to reach an optimal point that maximizes inter-cluster separability. We conducted experiments on both synthetic and real datasets, showing that the proposed bi-level optimization framework can effectively learn clustering assignments together with meaningful unsupervised representations. We also observed some limitations in the current approach. In particular, the dimension reduction component does not yet leverage deep network architectures. Therefore, we evaluated the method under relatively shallow network settings. As discussed, the proposed framework is general and can be extended to deeper neural networks, which represents a promising direction for future work.

\bibliographystyle{IEEEtran}
\bibliography{egbib}

\end{document}